\documentclass[sigconf]{acmart}
\renewcommand\footnotetextcopyrightpermission[1]{} 
\AtBeginDocument{%
  \providecommand\BibTeX{{%
    \normalfont B\kern-0.5em{\scshape i\kern-0.25em b}\kern-0.8em\TeX}}}

\acmConference[]{}{}{}

\usepackage{cite}
\usepackage{float}
\usepackage{amsmath,amssymb,amsfonts}
\usepackage{algorithmic}
\usepackage{graphicx}
\usepackage{textcomp}
\usepackage{xcolor}
\def\BibTeX{{\rm B\kern-.05em{\sc i\kern-.025em b}\kern-.08em
    T\kern-.1667em\lower.7ex\hbox{E}\kern-.125emX}}
    
\usepackage{graphicx}
\usepackage{times}
\usepackage{soul}
\usepackage{url}
\usepackage[utf8]{inputenc}
\usepackage{graphicx}
\usepackage{amsmath}
\usepackage{amsthm}
\usepackage{booktabs}
\usepackage{algorithm}
\usepackage{algorithmic}
\usepackage{makecell}
\usepackage{xcolor}
\usepackage{graphicx}
\usepackage{bm}
\usepackage{bbm}
\usepackage{array}
\usepackage{subfigure}
\usepackage{tikz}
\usepackage{amssymb}
\usepackage{verbatim}
\usepackage{enumitem}
\usepackage{booktabs}
\usepackage{dsfont}
\usepackage{enumitem}
\usepackage{mathtools}

\begin{document}

\title{Compact Multi-Head Self-Attention for Learning Supervised Text Representations}

\author{Sneha Mehta}
\affiliation{%
  \institution{Virginia Tech, USA}
}
\email{snehamehta@cs.vt.edu}

\author{Huzefa Rangwala}
\affiliation{%
  \institution{George Mason University, USA}
  }
\email{rangwala@cs.gmu.edu}

\author{Naren Ramakrishnan}
\affiliation{
  \institution{Virginia Tech, USA}
}
\email{naren@cs.vt.edu}


\begin{abstract}
Effective representation learning from text has been an active area of research in the fields of NLP and text mining. Attention mechanisms have been at the forefront in order to learn contextual sentence representations. Current state-of-the-art approaches for many NLP tasks use large pre-trained language models such as BERT, XLNet and so on for learning representations. These models are based on the Transformer architecture that involves recurrent blocks of computation consisting of multi-head self-attention and feedforward networks.  One of the major bottlenecks largely contributing to the computational complexity of the Transformer models is the self-attention layer, that is both computationally expensive and parameter intensive. 
In this work, we introduce a novel multi-head self-attention mechanism operating on GRUs that is shown to be computationally cheaper and more parameter efficient than self-attention mechanism proposed in Transformers for text classification tasks. The efficiency of our approach mainly stems from two optimizations; 1) we use low-rank matrix factorization of the affinity matrix to efficiently get multiple attention distributions instead of having separate parameters for each head
2) attention scores are obtained by querying a global context vector instead of densely querying all the words in the sentence. We evaluate the performance of the proposed model on tasks such as sentiment analysis from movie reviews, predicting business ratings from reviews and classifying news articles into topics. We find that the proposed approach matches or outperforms a series of strong baselines and is more parameter efficient than comparable multi-head approaches. We also perform qualitative analyses to verify that the proposed approach is interpretable and captures context-dependent word importance.
\end{abstract}





\maketitle

\section{Introduction}
Learning effective language representation is important for a variety of text analysis tasks including sentiment analysis, news classification, natural language inference and question answering. 
Supervised learning using neural networks commonly entails learning intermediate sentence representations followed by a task specific layer. For text classification tasks; this is usually a fully connected layer followed by an \textit{N-way} softmax where \textit{N} is the number of classes. 

Learning self-supervised language representations has made substantial progress in recent years with the introduction of new techniques for language modeling  combined with deep models like ELMo~\citep{peters2018deep}, ULMFit ~\citep{howard2018universal} and more recently BERT ~\citep{devlin2018bert} and GPT-2 ~\citep{radford2019language}. There has been a surge of BERT-based language models that use larger data for pretraining and different pretraining methods such as XL-Net ~\citep{yang2019xlnet}, RoBERTa ~\citep{liu2019roberta} while others combine different modalities ~\citep{su2019vl}. These methods have enabled transfer of learned representations via pre-training to downstream tasks.
Although these models work well on a variety of tasks there are two major limitations: 1) they are  computationally expensive to train 2) they usually have a large number of parameters that greatly increases the model size and memory requirements. For instance, the multilingual BERT-base cased model has 110M parameters, the small GPT-2 model has 117M parameters ~\citep{radford2019language} and the RoBERTa model was trained on 160GB of data ~\citep{liu2019roberta}. Recently, researchers have proposed `lighter' BERT models that leverage knowledge distillation during the pre-training phase and reduce the size of the BERT model ~\citep{wolf2019transformers} or  try to improve the parameter efficiency of the BERT model by optimizations at the embedding layer and by parameter sharing ~\citep{lan2019albert}. However, all of the above models are based on the Transformer architecture ~\citep{vaswani2017attention} the major component of which is the scaled dot-product self-attention mechanism .   

This layer has a computational complexity of $O(n^2*d)$ that scales quadratically with the length of the input ($n$) and linearly with the length of the model hidden size ($d)$.
It is natural to see how task specific training or fine-tuning can be limiting when the  training data and computational resources are scarce and sequences are long. 
Further, running inference on and storing such models can also be difficult in low resource scenarios such as IoT devices or low-latency use cases. Hence, supervised learning for task-specific architectures which are trained from scratch, especially where domain specific training data is available are useful. They are light-weight and easy to deploy. In this work, we propose \textbf{l}ow-rank f\textbf{a}ctorization based \textbf{m}ulti-head \textbf{a}ttention mechanism (LAMA), a lean attention mechanism which is computationally cheaper and more parameter efficient than prior approaches and exceeds or matches the performance of state-of-the-art baselines including large pretrained models, with fewer parameters.
Contrary to previous approaches ~\citep{lin2017structured,guo2017end} that are based on additive attention mechanism ~\citep{neuralAlign}, LAMA is based on multiplicative attention ~\citep{luong2015effective} which replaced the additive attention by dot product for faster computation. We further introduce a bilinear projection while computing the dot product to capture similarities between a global context vector and each word in the sentence ~\citep{luong2015effective}. The function of the bilinear projection is to capture nuanced context dependent word-importance as corroborated by previous works ~\citep{chen2017reading}. Next, unlike previous methods we use low-rank formulation of the bilinear projection matrix based on hadarmard product ~\citep{kim2016hadamard,yu2015combining}
to produce multiple attentions by querying the global context vector for each word as opposed to having a different learned vector ~\citep{lin2017structured} or matrix ~\citep{yu2017multi}. In effect, we cast score computation between a word representation and a context vector as unimodal feature fusion to a low dimensional space akin to its multimodal counterparts ~\citep{yu2017multi,lin2015bilinear}. Each dimension of this low-dimensional feature space can be considered as the contribution of the word to a different attention head. By controlling the dimension of this feature space new heads can be added or removed. Finally, we devise a mechanism to obtain context-aware supervised sentence embeddings from the obtained attention distributions for downstream classification tasks. We evaluate our model on tasks such as sentiment analysis, predicting business ratings and news classification. We show that the proposed model learns context-dependent word importance like other attention models. Moreover, the proposed model is 3 times more parameter efficient than other comparable attention models especially the encoder of a Transformer model ~\citep{vaswani2017attention}. In terms of performance, the proposed model is competitive and matches or exceeds the performance of strong baselines, attention and non-attention based. Further, we present some ancillary analyses on model efficiency and need for multiple attention heads. In summary, our results show that the proposed model can be reliably used as a leaner multi-head attention alternative for supervised text classification tasks.

The organization of the rest of the paper is as follows: the next section (\S\ref{sec:rw}) discusses connections with related work, followed by the description of the proposed model (\S\ref{sec:model}), followed by the task descriptions and  the experimental evaluation (\S\ref{sec:experiments}). Finally, in the following sections the results are presented along with their discussion (\S\ref{sec:results}), before concluding (\S\ref{sec:conclusion}). 

\section{Related Work} \label{sec:rw}
Spearheaded by their success in neural machine translation ~\citep{neuralAlign,luong2015effective} attention mechanisms are now ubiquitously used in problems such as question answering ~\citep{teachQA, seo2016bidirectional, dhingra2016gated}, text summarization ~\citep{paulus2017deep, chen2017reading}, event extraction ~\citep{Mehta:2019:EDU:3308558.3313659}, and training large language models ~\citep{devlin2018bert,radford2019language}. In sequence modeling, attention mechanisms allow the decoder to learn which part of the sequence it should ``attend'' to based on the input sequence and the output it has generated so far ~\citep{neuralAlign}.

\subsection{Self-Attention}
Self-attention, sometimes called intra-attention is an attention mechanism relating different positions of a single sequence in order to compute a representation of the sequence ~\citep{cheng2016long}. Self-attention has been
used successfully in a variety of tasks including reading comprehension, abstractive summarization, textual entailment and learning task-independent sentence representations ~\citep{cheng2016long, parikh2016decomposable, paulus2017deep,lin2017structured}.
Traditionally, the above methods have depended on Recurrent Neural Networks(RNNs)~\citep{Hochreiter:1997:LSM:1246443.1246450,Chung2014EmpiricalEO} to model sequential dependencies and attention mechanisms were proposed to alleviate the vanishing gradient problem by establishing shorter connections between the source and target positions. The inherently
sequential nature of RNNs precludes parallelization within training examples which becomes a memory bottleneck for batching across examples. Recently Transformer model was proposed ~\citep{vaswani2017attention} that replaced the dependence on RNNs and relies completely on the self-attention mechanism. In this approach, every position attends to every other position and adjusts its embedding accordingly. However, this dense computation is quadratic in the length of the input and is resource intensive. 
The proposed approach on the contrary leverages RNNs for modeling sequential dependencies and uses a single global query vector for each input word. We show that our approach is computationally more efficient in comparison to encoder of the Transformer model on text classification tasks and exceeds in performance.

\subsection{Multi-Head Attention}
Models have been proposed that compute multiple attention distributions over a single sequence of words. Multi-view networks ~\citep{guo2017end} use a different set of parameters for each view which leads to an increase in the number of parameters.
Lin et al. ~\citep{lin2017structured} use the additive attention mechanism, which is a more general approach of computing attention between query and key vectors and modify it to produce multiple attentions to obtain a matrix sentence embedding.
 Scaled dot product attention proposed by Vaswani et. al. ~\citep{vaswani2017attention} is a direct approach and is based on dot product between the query and key vectors. This approach has been shown to be very effective in machine translation ~\citep{vaswani2017attention} and pretraining language models ~\citep{devlin2018bert}. However to compute multiple attention heads different transformation parameters are learned for different heads leading to increased parameters.
In this work, on the other hand, score between the key (context) and the query (word representation) is computed using a bilinear projection matrix followed by an approach inspired by multi-modal low rank bilinear pooling ~\citep{kim2016hadamard} to factorize the matrix into two low rank matrices to compute multiple attention distributions over words.  We find that this is a more parameter efficient way of computing multiple attentions. Contrary to Guo et al. ~\citep{guo2017end} and Vaswani et al. ~\citep{vaswani2017attention} we use matrix factorization to alleviate the problem of increasing parameters with increasing heads and the proposed model performs superior to their approach.

\subsection{Low-Rank Factorization}
Low-rank factorization has been a popular approach to reduce the size of the hidden layers ~\citep{chen2018adaptive,tai2015convolutional}. Recent work has achieved
significant improvements in computational efficiency through factorization tricks ~\citep{shazeer2017outrageously} and conditional
computation ~\citep{kuchaiev2017factorization}. Recently, factorization was also employed at the embedding layer of large pretrained language models such as BERT ~\citep{devlin2018bert} to reduce the model size ~\citep{lan2019albert}. In this work, we employ factorization for unimodal feature fusion for computing attention scores for multiple attention heads. Hadamard product formulation for matrix factorization is used to compactify the multi-head attention layer. This formulation can be viewed as low-rank bilinear pooling for unimodal features based on the corresponding idea of multimodal feature fusion ~\citep{yu2017multi,lin2015bilinear}.



\section{Proposed Model} \label{sec:model}
In this section, we give an overview of the proposed model followed by a detailed description of each model component.

A document (a product review or a news article) is first tokenized and converted to a word embedding via a lookup into a pretrained embedding matrix. The embedding of each token is encoded via a bi-directional Gated Recurrent Unit ~\citep{cho2014learning} (bi-GRU) sentence encoder to get a contextual annotation of each word in that document. The LAMA attention mechanism then obtains multiple attention distributions over those words by computing an alignment score of their hidden representation with a word-level context vector. Sum of the word representations weighted by the scores from multiple attention distributions then forms a matrix document embedding. The matrix embedding is then flattened and passed onto downstream layers (either a classifier or another encoder depending on the task). 

In the rest of the paper, capital bold letters indicate matrices, small bold letters indicate vectors and small letters indicate scalars.
 
 
\subsection{Sequence Encoder}
 We use the GRU ~\citep{neuralAlign} RNN as the sequence encoder. GRU uses a gating mechanism to track the state of the sequences. There are two types of gates: the reset gate $\textbf{r}_t$ and the update gate $\textbf{z}_t$. The update gate decides how much past information is kept and how much new information is added.  At time $t$, the GRU computes its new state as:
\begin{align}
 \textbf{h}_t = (1-\textbf{z}_t) \odot \textbf{h}_{t-1} + \textbf{z}_t  \odot \Tilde{\textbf{h}}
\end{align}
and the update gate $\textbf{z}_t$ is updated as:
 \begin{align}
   \textbf{z}_t = \sigma(\textbf{W}_z*\textbf{x}_t + \textbf{U}_z*\textbf{h}_{t-1} + \textbf{b}_z)   
 \end{align}
The RNN candidate state $\Tilde{\textbf{h}_t}$ is computed as:
\begin{align}
\Tilde{\textbf{h}_t} = tanh(\textbf{W}_h\textbf{x}_t + \textbf{r}_t \odot (\textbf{U}_h \textbf{h}_{t-1} + \textbf{b}_h)
\end{align}
Here $\textbf{r}_t$ is the reset gate which controls how much the past state contributes to the candidate state. If $\textbf{r}_t$ is zero, then it forgets the previous state. The reset gate is updated as follows:
\begin{align}
\textbf{r}_t = \sigma(\textbf{W}_r\textbf{x}_t + \textbf{U}_r\textbf{h}_{t-1} + \textbf{b}_r)
\end{align}
Consider a document $D_i$ containing $T$ words. $D_i=\{\textbf{w}_{1},...,\textbf{w}_{t},...,\textbf{w}_{T}\}$. Let each word be denoted by $\textbf{w}_{t}$, $t \in [0, T]$  where every word is converted to a real valued word vector $\textbf{x}_{t}$ using a pre-trained embedding matrix $\textbf{W}_e = R^{d \times |V|}$, $\textbf{x}_{t} = \textbf{W}_e \textbf{w}_{t}$, $t \in [1, T]$  where $d$ is the embedding dimension and $V$ is the vocabulary. The embedding matrix $\textbf{W}_e$ is fine-tuned during training. Note that we have dropped the subscript $i$ as all the derivations are for the $i^{th}$ document and it is assumed implicit in the following sections unless otherwise stated.

We encode the document using a bi-GRU that summarizes information in both directions along the text to get a contextual annotation of a word. In a bi-GRU the hidden state at time step $t$ is represented as a concatenation of hidden states in the forward and backward direction. The forward GRU denoted by $\overrightarrow{GRU}$ processes the document from $w_{1}$ to $w_{T}$ whereas the backward GRU denoted by $\overleftarrow{GRU}$ processes it from $w_{T}$ to $w_{1}$.

\begin{align}
\textbf{x}_{t} = \textbf{W}_e \textbf{w}_{t}
\end{align}
\begin{subequations}
\begin{align}
\overrightarrow{\textbf{h}_{t}} = \overrightarrow{GRU}(\textbf{x}_{t}, \textbf{h}_{(t-1)}, \mbox{\boldmath$\theta$})
\end{align}
\begin{align}
\overleftarrow{\textbf{h}_{t}} = \overleftarrow{GRU}(\textbf{x}_{t}, \textbf{h}_{(t+1)}, \mbox{\boldmath$\theta$})
\end{align}
\end{subequations}
Here the word annotation $\textbf{h}_{t}$ is obtained by concatenating the forward hidden state $\overrightarrow{\textbf{h}_{t}}$ and the backward hidden state $\overleftarrow{\textbf{h}_{t}}$.

\subsection{Single-Head Attention}
To alleviate the burden of remembering long term dependencies from GRUs we use the global attention mechanism ~\citep{luong2015effective} in which the document representation is computed by attending to all words in the document. Let $\textbf{h}_{t}$ be the annotation corresponding to the word $\textbf{x}_{t}$. First we transform $\textbf{h}_{t}$ using a one layer Multi-Layer Perceptron (MLP) to obtain its hidden representation $\textbf{u}_{t}$. We assume Gaussian priors with $0$ mean and $0.1$ standard deviation on $\textbf{W}_w$ and $\textbf{b}_w$.
\begin{align} \label{eq:affine_word}
\textbf{u}_{t} = tanh(\textbf{W}_w \textbf{h}_{t} + \textbf{b}_w)
\end{align}
Next, to compute the importance of the word in the current context we calculate its relevance to a global context vector $\textbf{c}$ using a bilinear projection. 
\begin{align}
 f_{t} =  \textbf{c}^\top \textbf{W}_i \textbf{u}_{t} \label{eq:bil}
\end{align}
Here, $\textbf{W}_i \in {\rm I\!R}^{2h \times 2h}$, is a bilinear projection matrix which is randomly initialized and jointly learned with other parameters during training. $h$ is the dimension of the GRU hidden state and $\textbf{u}_t$ \& $\textbf{c}$ are both of dimension $2h \times  1$ since we're using a bi-GRU. The mean of the word embeddings provides a good initial approximation of the global context of the document. We initialize $\textbf{c} = \frac{1}{T}\sum_{t=1}^T \textbf{w}_t$ which is then updated during training. 
We use a bilinear projection because they are more effective in learning pairwise interactions as shown in previous works ~\citep{chen2017reading}.  The attention weight for the word $\textbf{x}_{t}$ is then computed using a \textit{softmax} function where summation is taken over all the words in the document.
\begin{align} \label{eq:single_head}
   \alpha_{t} = \frac{exp(f_{t})}{\sum_{t'} exp(f_{t'})}
\end{align}

\begin{table}[]
\caption{Important notations. Capital bold letters indicate matrices, small bold letters indicate vectors, small letters indicate scalars.}
\label{tab:note}
\small
\begin{center}
\begin{tabular}{|l|l|} \hline
\textbf{Notation} & \textbf{Meaning}  \\ \hline
 $N$ & Corpus size \\ 
 $T$ & \# of words tokens in a sample \\
 $m$ & \# of attention heads   \\
 $f_{t}$ & alignment score \\
 $\alpha_{t}$ & attention weight \\
 $\textbf{u}_{t}$ & word hidden representation \\
 $\textbf{c}$ & global context vector \\
 $h$ & GRU hidden state dimension \\
\hline
\end{tabular}
\end{center}
\vspace{-1em}
\end{table}

\begin{figure*}[ht!]
 \centering
 \small
  \includegraphics[width=\textwidth]{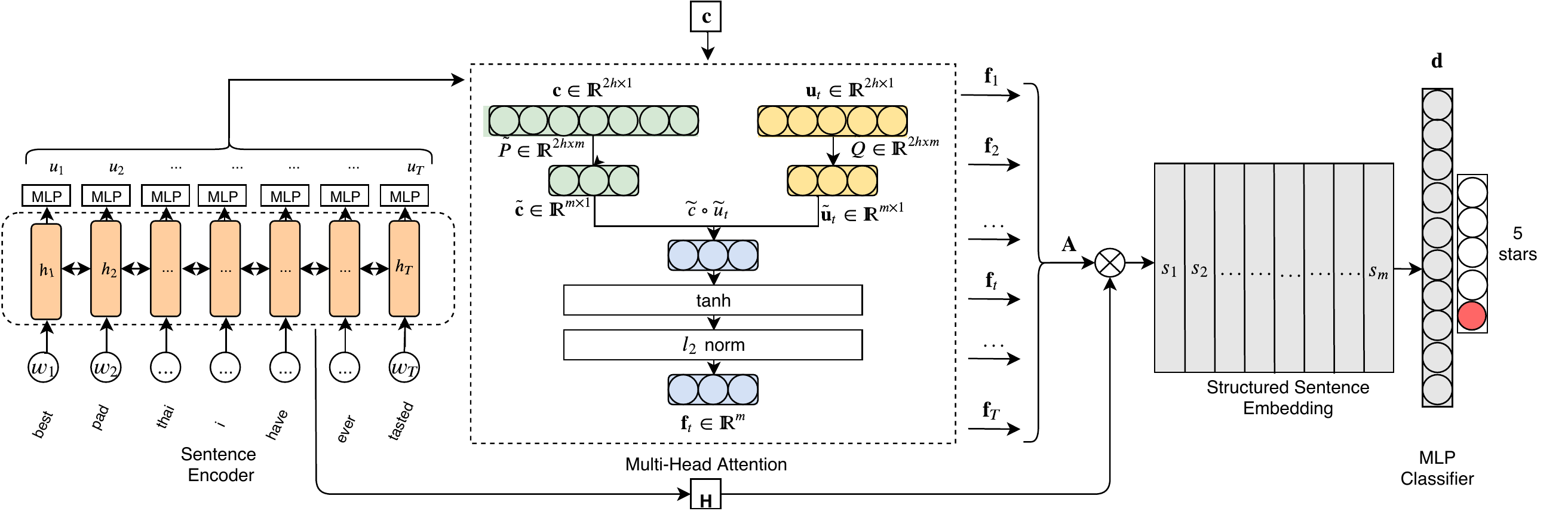}
 \caption{Figure shows a schematic of the model architecture and its major components including the Sentence Encoder, proposed multi-head attention mechanism LAMA, Structured Sentence Embedding and finally the MLP classifier. The attention computation is demonstrated for a single word.}
 \label{fig:desc}
\vspace{-1em}
\end{figure*}

\subsection{Low-Rank Factorization based Multi-Head Attention}
\label{sec:lama}
In this section, we describe the novel low-rank factorization based multi-head attention mechanism (LAMA).
The attention distribution in Eq. ~\ref{eq:single_head} above usually focuses on a specific component of the document, like a special set of trigger words. So it is expected to reflect an aspect, or component of the semantics in a document. This type of attention is useful for smaller pieces of texts such as tweets or short reviews. For larger reviews there can be multiple aspects that describe that review.  For this we introduce a novel way of computing multiple heads of attention that capture different aspects.

Suppose $m$ heads of attention are to be computed, we need $m$ alignment scores between each word hidden representation $\textbf{u}_{t}$ and the context vector $\textbf{c}$. To obtain an $m$ dimensional output $\textbf{f}_{t}$, we need to learn $m$ weight matrices given by $\textbf{W} = [\textbf{W}_1, ...,\textbf{W}_m] \in \mathbf{R}^{m \times 2h \times 2h}$ as demonstrated in previous works. Although this strategy might be effective in capturing pairwise interactions for each aspect it also introduces a huge number of parameters that may lead to overfitting and also incur a high computational cost especially for a large $m$ or a large $h$. To address this, the rank of matrix $\textbf{W}$ can be reduced by using low-rank bilinear method to have less number of parameters ~\citep{kim2016hadamard,yu2017multi}. Consider one head; the bilinear projection matrix $\textbf{W}_i$ in Eq. \ref{eq:bil} can be factorized into two low rank matrices $\textbf{P}$ \& $\textbf{Q}$.

\begin{align}
 f_{t} = \textbf{c}^\top  \textbf{P}  \textbf{Q}^\top \textbf{u}_{t} = \sum_{d=1}^k \textbf{c}^\top p_d q_d^\top \textbf{u}_{t} 
 =   \mathbbm{1}^\top(\textbf{P}^\top \textbf{c} \circ \textbf{Q}^\top \textbf{u}_{t}) \label{eq:decompose}
\end{align}
where $\textbf{P} = [p_{1}, ..., p_{k} ]  \in {\rm I\!R}^{2h \times k}$ and $\textbf{Q} = [q_{1}, ..., q_{k}] \in {\rm I\!R}^{2h \times k}$ are two low-rank matrics,  $\circ$ is the Hadamard product or the element-wise multiplication of two vectors, $\mathbbm{1} \in {\rm I\!R}^k$ is an all-one vector and $k$ is the latent dimensionality of the factorized matrices.

To obtain $m$ scores, by Eq.\ref{eq:decompose}, the weights to be learned are two three-order tensors
$\mathds{P} = [\textbf{P}_{1}, ..., \textbf{P}_{m}] \in {\rm I\!R}^{2h \times k \times m}$ and $\mathds{Q} = [\textbf{Q}_{1}, ..., \textbf{Q}_{m}] \in {\rm I\!R}^{2h \times k \times m}$
accordingly. Without loss of generality 
$\mathds{P}$ and $\mathds{Q}$ can be reformulated as 2-D matrices $\Tilde{\mathds{P}} \in  {\rm I\!R}^{2h\times km}$ and $\Tilde{\mathds{Q}} \in {\rm I\!R}^{2h \times km}$ respectively with simple reshape operations. Setting $k=1$, which corresponds to rank-1 factorization. Eq.\ref{eq:decompose} can be written as: 
\begin{align} \label{eq:lama}
  \textbf{f}_{t} = \Tilde{\mathds{P}}^\top\textbf{c} \circ \Tilde{\mathds{Q}}^\top\textbf{u}_{t}
\end{align}
This brings the two feature vectors $\textbf{u}_{t} \in \mathbf{R}^{2h}$, the word hidden representation and $\textbf{c} \in \mathbf{R}^{2h}$, the global context vector in a common subspace and are given by $\Tilde{\textbf{u}}_{t}$ and $\Tilde{\textbf{c}}$ respectively.
$\textbf{f}_{t} \in {\rm I\!R}^m$ can be viewed as a multi-head alignment vector for the word $\textbf{x}_{t}$ where each dimension of the vector can be viewed as the score of the word w.r.t. a different attention head. For computing attention for one head, this is equivalent to replacing the projection matrix \textbf{$\textbf{W}_i$} in Eq. ~\ref{eq:bil} by the outer product of vectors $\Tilde{\mathds{P}_i}$ and $\Tilde{\mathds{Q}_i}$;  rows of the matrices $\Tilde{\mathds{P}}$ and $\Tilde{\mathds{Q}}$ respectively and rewriting it as the Hadamard product. As a result each row of matrices $\Tilde{\mathds{P}_i}$ and $\Tilde{\mathds{Q}_i}$ represent the vectors for computing the score for a different head.

The multi-head attention vector $\bm{\alpha}_{t} \in {\rm I\!R}^m$ is obtained by computing a softmax function along the sentence length: 
\begin{align}
  \bm{\alpha}_{t} =  \frac{exp(\textbf{f}_{t})}{\sum_{t'} exp(\textbf{f}_{t'})}
\end{align}
Before computing softmax, similar to previous works ~\citep{kim2016hadamard,yu2017multi}, to further increase the model capacity we apply the \textit{tanh} nonlinearlity to $\textbf{f}_{t}$. Since element-wise multiplication is introduced the variance of the model might increase, so we apply an $l_2$ normalization layer across the $m$ dimension. Although $l_2$ is not strictly necessary since both $\textbf{c}$ and $\textbf{u}_t$ are in the same modality empirically we do see improvement after applying $l_2$. Each component $k$ of $\bm{\alpha}_{t}$ is the contribution of the word $\textbf{x}_t$ to the $k^{th}$ head.  

Next, we describe how this computation can be vectorized for each word in the document. Let $\textbf{H} = (\textbf{h}_{1}, \textbf{h}_{2},...\textbf{h}_{T})$ be a matrix of all word annotations in the sentence; $\textbf{H} \in {\rm I\!R}^{T \times 2h}$.
The attention matrix for the sentence can be computed as: 
\begin{align} 
\textbf{A} = softmax(l2(tanh(\tilde{\mathds{P}}^\top \textbf{C}_g\circ \tilde{\mathds{Q}}^\top  \textbf{H}^\top)))  \label{eq:attn}
\end{align}
where, $\textbf{C}_g \in  {\rm I\!R}^{2h \times T}$ is $\textbf{c}$ repeated $T$ times, once for each word, $l_2(\textbf{x}) = \frac{\textbf{x}}{||\textbf{x}||}$ and \textit{softmax} is applied row-wise. $\textbf{A} \in {\rm I\!R}^{m \times T}$ is the attention matrix between the sentence and the global context with each row representing attention for one aspect.


Given $\textbf{A} = [\bm{\alpha}_{1}, \bm{\alpha}_{2},...\bm{\alpha}_{T}]$, the multi-head attention matrix for the sentence; $\textbf{A} \in {\rm I\!R}^{m \times T}$. The document representation for a head $j$ given by $\bm{\alpha}_{j} = \{\alpha_{j1}, \alpha_{j2},..\alpha_{jT}\}$ can be computed by taking a weighted sum of all word annotations.
\begin{align}
\textbf{s}_{j} = \sum_{k=1}^T \textbf{h}_{k}*\alpha_{jk}
\end{align}
Similarly, document representation can be computed for all heads and is given in a compact form by:
\begin{align}
    \textbf{S} = \textbf{A}\textbf{H} \label{eq:sent_emb}
\end{align}
Here $\textbf{S} \in {\rm I\!R}^{m\times2h}$ is a matrix sentence embedding and contains as many rows as the number of heads. Each row contains an attention distribution for a new head. It is flattened by concatenating all rows to obtain the document representation $\textbf{d}$. From the document representation, the class probabilities are obtained as follows.
\begin{align}
\hat{\textbf{y}} = softmax(\textbf{W}_c \textbf{d} + \textbf{b}_c)
\end{align}
Loss is computed using cross entropy. 
\begin{align}
L(\textbf{y}, \hat{\textbf{y}}) = -\sum_{c=1}^C y_c log(\hat{y_c})
\end{align}
where $C$ is the number of classes and $\hat{y_c}$ is the probability of the class $c$.

\subsection{Disagreement Regularization}
To reduce the variance of model introduced due to point-wise multiplication and to encourage diversity among multiple attention heads we introduce an auxiliary regularization term.
\begin{equation}
    J(\theta) = arg min_{\theta} \{ L(\textbf{y}, \hat{\textbf{y}}) - \lambda * D(\textbf{A} | \textbf{x}, \textbf{y}; \theta) \} \label{eq:loss}
\end{equation}
where \textbf{A} is the attention matrix, $\theta$ represents model parameters, $\lambda$ is a hyper-parameter and is empirically set to 0.2
in this paper. $L(\hat{\textbf{y}}, \textbf{y})$ is the cross-entropy loss and $D(\textbf{A} | \textbf{x}, \textbf{y}; \theta)$ is the auxiliary regularization term that represents the disagreement between different attentions. It guides the related attention components to capture different features from the corresponding projected subspaces.
We try two different regularizations i) regularization over attended positions ii) regularization over document embeddings. For the first type we adapt the penalization term in ~\citep{lin2017structured} to represent disagreement between attention distributions (Eq. \ref{eq:dpenal}).
Next, we directly regularize the sentence embeddings resulting from different attention distributions represented using their cosine similarity (Eq. \ref{eq:demb}). The more similar the embeddings, lesser the disagreement.
\begin{subequations}
\begin{align}
     D_{penal} =  - \lVert \textbf{A}\textbf{A}^\top - \textbf{I} \rVert^2_F  \label{eq:dpenal} 
\end{align}
\begin{align}
    D_{emb} = -\frac{1}{m^2} \sum_{i=1}^m \sum_{i=1}^m \frac{\textbf{s}_i \cdot \textbf{s}_j}{\lVert \textbf{s}_i \rVert \lVert \textbf{s}_j \rVert} \label{eq:demb}
\end{align}
\end{subequations}
The final training loss is given by Eq. \ref{eq:loss}
summed over all documents in a minibatch.
We use the minibatch stochastic gradient descent algorithm ~\citep{kiefer1952stochastic} with momentum and weight decay for optimizing the loss function and the backpropogation algorithm is used to compute the gradients. 

Fig. \ref{fig:desc} shows a single document and its flow through various model components. The middle block illustrates the proposed attention mechanism for one word $\textbf{w}_t$ of the document. It is first transformed through Eq. ~\ref{eq:affine_word} to $\textbf{u}_t$. In parallel, a context vector $\textbf{c}$ is initialized. Eq. ~\ref{eq:lama} is then used to compute the multi-head attention for this word. Vectorized attention computation can be performed for all words using Eq. ~\ref{eq:attn}  to obtain the attention matrix $\textbf{A}$ which is then multiplied by the hidden state matrix $\textbf{H}$ to obtain an embedding for the document which is then passed to the MLP classifier.

\subsection{Hyperparameters}
We use a word embedding size of 100. The embedding matrix $\textbf{W}_e$ is pretrained on the corpus using word2vec ~\citep{mikolov2013distributed}. All words appearing less than 5 times are discarded. The GRU hidden state is set to $h=50$, MLP hidden state to $512$ and we apply a dropout of $0.4$ to the hidden layer. Batch size is set to $32$ for training and an initial learning rate of $0.05$ is used. For early stopping we use $patience=5$. Momentum is set to 0.9 and weight decay to 0.0001. 
We will open source the code on acceptance. 

\subsection{Computational Complexity}
Other variables such as document encoder and dimension of the hidden state held constant, the computational complexity of the model depends on the attention layer. It can be seen that the computational complexity of LAMA is linear in input and linear in the number of attention heads. Where as, the attention mechanism in the encoder of a Transformer model ~\citep{vaswani2017attention} is quadratic in the length of the input (Table ~\ref{tab:complexity}). For cases where $n*m \ll n^2$, which is a common scenario, attention mechanism LAMA is computationally more efficient than self-attention in encoder of the Transformer model for sequence classification tasks.

\begin{table}[]
\centering
\caption{\small{Per-layer complexity for different layer types. $n$ is the document length, $m$ is the number of attention heads and $d$ is the representation dimension.}}
\label{tab:complexity}
\begin{tabular}{@{}ll@{}}
\toprule
Layer Type          & Complexity per Layer \\ \midrule
Self Attention (LAMA)               &    $O(n*m*h)$                  \\
Self-Attention (TE) &   $O(n^2*d)$                \\
\bottomrule
\end{tabular}
\end{table}
\section{Evaluation} \label{sec:experiments}

\subsection{Datasets}
We evaluate the performance of the proposed model on tasks of predicting business ratings from Yelp, sentiment prediction from movie reviews and classifying news articles into topics.  Table ~\ref{tab:dataset} gives an overview of the datasets and their statistics. 

\begin{table}[]
\centering
\caption{Dataset statistics. \# words indicates the average number of tokens per document in the corresponding datasets.}
  \label{tab:dataset}
\begin{tabular}{lllll}
Dataset & \# Classes & \# Train & \# Test & \# words \\ \hline
YELP & 5 & 499,976 &  4,000 & 118  \\
YELP-L & 5 & 175,844 & 1,378 & 226 \\
YELP-P & 2 & 560,000 & 38,000 & 137 \\ 
IMDB & 2 & 25,000  & 25,000 & 221 \\
Reuters & 8 & 4,484 & 2,189 & 102 \\
News & 4 & 151,328  & 32,428  & 352  \\ \hline
\end{tabular}
\end{table}


\subsubsection{Yelp}
The Yelp dataset ~\footnote{https://www.yelp.com/dataset challenge}  consists of 2.7M Yelp reviews and user ratings from 1 to 5. Given a review the goal is to predict the rating assigned by the user to the corresponding business store. We treat the task as 5-way text classification where each class indicates the user rating. We randomly sampled 500K review-star pairs as training set and 4,000 for test set. Reviews were tokenized using the Spacy tokenizer ~\footnote{https://spacy.io/}. 100-dimensional word embeddings were trained from scratch on the train dataset using the gensim ~\footnote{https://radimrehurek.com/gensim/} software package. 
\subsubsection{Yelp-Long}
Multi-head attention capturing multiple aspects is more useful for classifying ratings that are more subjective i.e. longer reviews where people express their experiences in detail. We create a subset of the YELP dataset containing all longer reviews i.e. reviews containing longer than 118 tokens which we found to be the mean length of the reviews in the dataset. The training set consists of 175,844  reviews, and the test set consists of 1,378 reviews. The goal is to predict the ratings from the above subset of the Yelp dataset. We refer to this dataset as Yelp-L (Yelp-Long) in the rest of the paper since it consists of all longer reviews. We hypothesize that having multi-head attention would benefit in this setting where more intricate foraging of information from different parts of the text is required to make a prediction.
The model hyperparameters and training settings remain the same as the above.

\subsubsection{Yelp-Polarity}
The Yelp reviews polarity dataset ~\citep{zhang2015character} is constructed by considering stars 1 and 2 negative, and 3 and 4 positive from the Yelp dataset. For each polarity 280,000 training samples and 19,000 testing samples are taken randomly. In total there are 560,000 training samples and 38,000 testing samples. This datset is referred as Yelp-P in the paper.


\subsubsection{Movie Reviews}
The large Movie Review dataset ~\citep{maas-EtAl:2011:ACL-HLT2011} contains movie reviews along with their associated binary
sentiment polarity labels. It contains 50,000 highly polar reviews ($score \leq 4$ out of 10 for negative reviews and $score >= 7$ out of 10 for positive reviews) split evenly into 25K train
and 25K test sets. The overall distribution of labels is balanced (25K pos and 25K neg).
In the entire collection, no more than 30 reviews are allowed for any given movie because reviews for the same movie tend to have correlated ratings. Further, the train and test sets contain a disjoint set of movies, so no significant performance is obtained by memorizing movie-unique terms and their associated with observed labels. 
We refer to this dataset as IMDB in the rest of the paper.

\subsubsection{News Aggregator} This dataset  ~\citep{Dua:2017} contains headlines, URLs, and categories for news stories collected by a web aggregator between March 10th, 2014 and August 10th, 2014. News categories included in this dataset include business; science and technology; entertainment; and health. Different news articles that refer to the same news item are also categorized together. Given a news article the task is to classify it into one of the four categories. Training dataset consits of 151,328 articles and test dataset consits of 32, 428. The average token length is 352.

\subsubsection{Reuters}
This dataset \footnote{https://www.cs.umb.edu/smimarog/textmining/datasets/} is taken from Reuters-21578 Text Categorization Collection. This is a collection of documents that appeared on Reuters newswire in 1987. The documents were assembled and indexed with categories. We evaluate on the Reuters8 dataset consisting of news articles about 8 topics including acq, crude, earn, grain, interest, money-fx, ship,trade.

\subsection{Comparative Methods}
To benchmark the proposed model against existing methods we use a variety of model architectures as our comparative baselines.
We use BERT ~\citep{devlin2018bert} as one of the baselines. In our experiments, we used the pretrained bert-base uncased model which has 12-layers, 768-hidden state size, 12-heads, 110M parameters and is trained on lower-cased English text. We finetuned it on our datasets for 2 epochs using the ADAM optimizer ~\citep{kingma2014adam} with a learning rate of $5e-6$. 

It has been shown that average of word embeddings can make for a very strong baseline ~\citep{shen2018baseline}. We use this as another baseline and refer to it as AVG in the paper.

We use a variety of models with and without attention as other baselines. Strong representative baselines for different model architectures are chosen; such as CNN with \textit{max-over-time} pooling ~\citep{kim2014convolutional}, bidirectional GRU ~\citep{Chung2014EmpiricalEO} model with maxpooling referred as BiGRU. We experimented with $n=1$ and $n=2$ GRU layers and found that $n=1$ converged faster and led to a better performance. For the CNN baseline we used 3 kernels of sizes 3, 4 and 5 with 100 filters each.

Among supervised attention-based multi-head models we use the Self Attention Network proposed by Lin et al. ~\citep{lin2017structured}. We refer to this baseline as SAN. For each task, we empirically find the number of heads that give the best performance.  We use a 1-layer Encoder of the Transformer model (TE) ~\citep{vaswani2017attention} with 8 attention heads as another baseline.  We use $d_{model}=512$ such that $d_{model}/\text{heads} = 64$. 

We use two variations of the proposed model to compare the performance with the above baselines. For the first variation we initialize $\textbf{c}$ with mean of  word embeddings in the sentence to provide the global context of the sentence. We call this baseline \textsf{LAMA+ctx}. In another variation, we randomly initialize $\textbf{c}$ and jointly learn it with other model parameters (\textsf{LAMA}). Our model with embedding regularization is referred as \textsf{LAMA+ctx}+$d_{e}$ and our model with regularization over positions is referred as \textsf{LAMA+ctx}+$d_{p}$. For the models \textsf{LAMA} and \textsf{LAMA+ctx} we empirically identify the optimal number of attention heads to get the best performance. 

\section{Results} \label{sec:results}

\begin{table*}[!t]
\caption{\small{Table reports the accuracy of the proposed models (\textsf{LAMA}, \textsf{LAMA+ctx}) against various baselines on sentiment analysis and news classification tasks. $+D_p$ refers to \textsf{LAMA + Ctx} with position-wise regularization whereas $+D_e$ refers to \textsf{LAMA + Ctx} with regularization over embeddings.  }}
\label{tab:performance}

\centering
\begin{tabular}{@{}l|lllllll@{}}
\toprule
Methods      & News  & Reuters & Yelp  & IMDB  & Yelp-L & Yelp-P  \\ 

\midrule 
SAN (Lin et al. ~\citep{lin2017structured})                 & 0.876 & 0.942   & 0.68  & 0.831 &  0.638 & 0.945 \\
BiGRU                 & 0.905 & 0.867   & 0.663 & 0.876 &  0.608 & 0.943    \\
CNN (Kim et al. ~\citep{kim2014convolutional})                 & 0.914 & 0.96    & 0.693 & 0.874 & \textbf{0.672}  & 0.953  \\
TE (Vaswani et al. ~\citep{vaswani2017attention})                 & 0.899 &  0.901  &  0.655& 0.817  &  0.569  & 0.925 \\
BERT (Devlin et al. ~\citep{devlin2018bert})                &    0.92   &   0.97     &  0.715     &    0.894   &    \textbf{0.672}  & \textbf{0.97}  \\
AVG (Arora et al. ~\citep{arora2017asimple})                & 0.91  & 0.795   & 0.653 & 0.874 &   0.652 & 0.928  \\
\textsf{LAMA}  & 0.922 & 0.965   & 0.697 & 0.895 &   0.653 & 0.947\\
\textsf{LAMA + Ctx}  & \textbf{0.923} & \textbf{0.973}   & \textbf{0.716} & \textbf{0.90}   & 0.665 & 0.952   \\

    + $D_{p}$&  0.903   & 0.948 & 0.711 & 0.874 & 0.656 & 0.948 \\
+ $D_{e} $ & 0.91 & 0.831  & 0.677 &  0.805 & 0.619 & 0.893 &   \\
 \bottomrule
\end{tabular}
\normalsize
\end{table*}

Table ~\ref{tab:performance} reports the accuracy of the best model on the test set after performing 3-fold cross-validation. 
The proposed model with global context initialization \textsf{LAMA+ctx} outperforms the SAN model ~\citep{lin2017structured} on all tasks from 3.3\% (Reuters) to 8.2\% (IMDB). This is due to the fact that during attention computation the proposed model architecture has the provision to access the global context of the sentence whereas for SAN no such provision is available.




Extrapolating the attention over larger chunks of text we get uniform attention over all words, which is equivalent to no attention or equal preference for all words. This is what a simple BiGRU effects to (in a contextual setting and average of word embeddings in a non-contextual setting). We note that \textsf{LAMA+ctx} outperforms BiGRU by 2.0\% (News), 12.2\% (Reuters) 7.9\% (Yelp) and 9.4 \% (Yelp-L) and 2.7 \% (IMDB). 

Our models outperform the Transformer Encoder (TE) on all tasks. It should be noted that this performance improvement also comes with fewer parameters than TE (Table ~\ref{tab:param_vs_m}). When compared to large fine-tuned pre-trained language models such as BERT we find that LAMA outperforms BERT on News, Reuters, Yelp and IMDB datasets. On YELP-L and YELP-P datasets BERT outperforms LAMA. However, it should be noted that besides being trained on large-scale text corpora and having a large memory footprint compared to LAMA, BERT models also take a longer time for pretraining. For instance, for Yelp-P it took 12.5 hours to train the model just for 1 epoch as compared to 20 mins for LAMA and for Yelp it took 8.5 hours as opposed to 25 mins for LAMA on one Nvidia Tesla P100 GPU.


For the non-contextual baseline of average of word embeddings there's an improvement of  1.4\% (News), 22.4\% (Reuters) 9.8\% (Yelp), 11.4\% (Yelp-L) and 3.0\% (IMDB)
which shows that contextual information captured by BiGRU or CNN models are indeed important for the tasks.

Compared to CNN models an improvement of 1\% (News), 1.4\% (Reuters) and 3.3\% (Yelp), and 3.0 \% (IMDB) can be observed.
On the Yelp-L dataset our model and CNN perform similarly. However, the proposed model is more interpretable and gives an option for inspecting the attended keywords.

Finally, our results suggest that using disagreement regularization on LAMA worsens the performance in general.

From the above results it can be noted that LAMA is a competitive, interpretable and lean supervised model for text classification tasks. 
\subsection{Parameter vs Heads}
 In Transformer-based models attention layer can prove to be a memory bottleneck when resources are constrained. In this section, we compare the number of trainable parameters of the proposed model (LAMA) and  Transformer Encoder (TE). Table ~\ref{tab:param_vs_m} shows the increase in number of parameters (y-axis) when the number of attention heads are varied as 2, 4, 8, 16, 32, 64. Powers of 2 are picked because TE requires number of attention heads to be divisble by $d_{model}$ ~\citep{vaswani2017attention}. Note that reported parameters also contain GRU \& embedding parameters for LAMA and feed forward layer \& position embedding parameters for TE, although these parameters don't depend on number of attention heads. To ensure a fair comparison we pick BiGRU hidden dimension and $d_{model}$ both as 512. The hidden layer dimension of the final layer sentence classifier is 1024, which is the same for both models. It can be observed that the number of parameters in TE is constant for all heads which is consistent with its definition, because with increasing heads; Key, Value and Query projections parameters are scaled down. For LAMA, the number of parameters increases only slightly with increasing number of heads because \textbf{P} \& \textbf{Q} are the only parameter matrices that are dependent on the number of attention heads $m$ and for which the size increases (linearly). More importantly, it should be noted that for the number of heads for most practical use cases, the proposed model LAMA is almost 3 times more compact than Transformer Encoder. 
 

\begin{table}[]
 \caption{Comparison of number of trainable parameters in the attention layer of LAMA and TE with increasing number of attention heads. Number of parameters increase linearly for LAMA whereas for TE they are constant. Overall LAMA is more parameter efficient than TE.}
 \centering
\begin{tabular}{@{}lll@{}}
\toprule
\# heads & \begin{tabular}[c]{@{}l@{}}\# parameters (LAMA)\\    (in millions)\end{tabular} & \begin{tabular}[c]{@{}l@{}}\# parameters (TE)\\    (in millions)\end{tabular} \\ \midrule
2        & 6.403                                                                           & 18.465                                                                        \\
4        & 6.405                                                                           & 18.465                                                                        \\
8        & 6.409                                                                           & 18.465                                                                        \\
16       & 6.418                                                                           & 18.465                                                                        \\
32       & 6.434                                                                           & 18.465                                                                        \\
64       & 6.468                                                                           & 18.465                                                                        \\
\bottomrule
\end{tabular}
 \label{tab:param_vs_m}
\end{table}


\subsection{Runtime} 
\begin{figure*}
 \centering
   \includegraphics[width=0.7\textwidth]{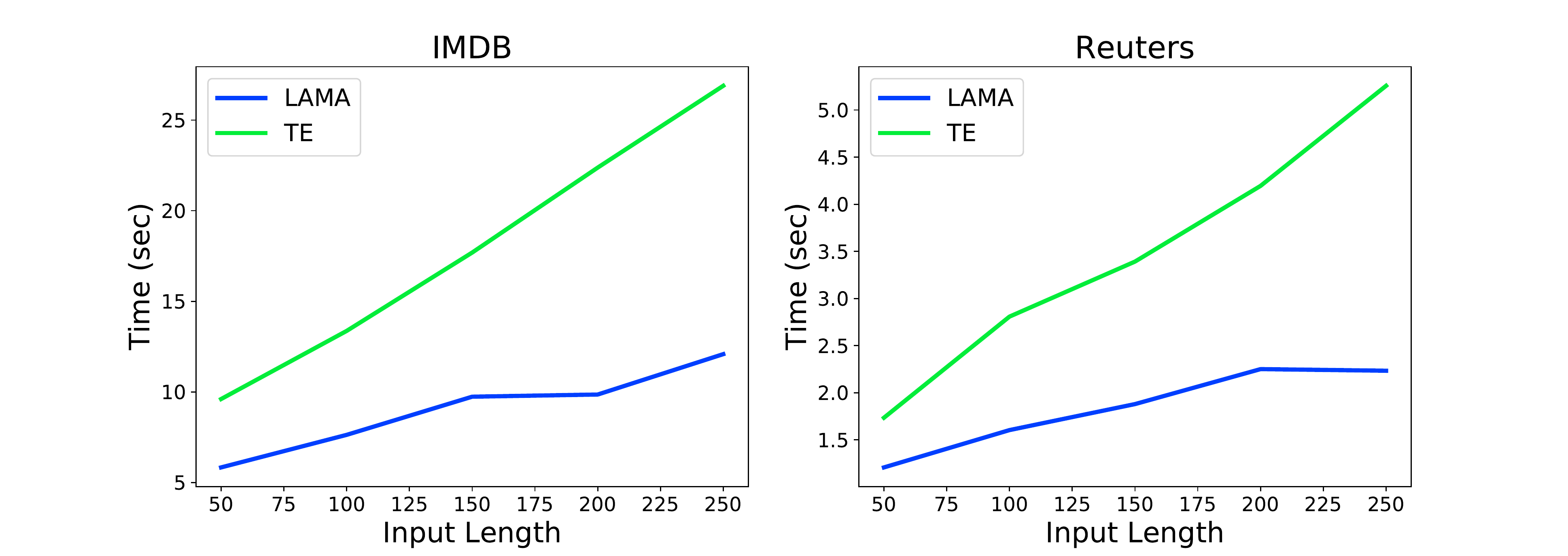}
 \caption{\small{Figure shows the average run time per epoch in seconds (averaged over 10 epochs) for LAMA Encoder (LE) and Transformer Encoder (TE) models as a function of input sequence length (50 to 250). It can be seen that TE (green) is more computationally expensive compared to LAMA (blue). LE is LAMA attention mechanism applied directly to word embeddings without computing GRU hidden states. } }
 \label{fig:time_vs_len}
 \end{figure*}

It should be noted that since LAMA is applied on top of RNN the computation is sequential and attention cannot be computed unless all RNN hidden states are available. Hence, the computational time increases linearly with input length and quadratically with hidden state dimension $(O(n*h^2))$. Once RNN hidden states are available the complexity of the attention layer LAMA is $O(n*m*h)$. In comparison, the complexity of the self-attention mechanism in Transformer is $O(n^2*d)$, where $d$ is the hidden state dimension. It increases quadratically with increasing input length and linearly with hidden state dimension. In this section, we compare the runtimes of LAMA and a 1-layer Transformer Encoder. For a fair comparison we compare the runtimes of LAMA free from RNN. We propose, LAMA Encoder -- a model that doesn't use RNN to model sequential dependencies and directly operates on the word embeddings of the inputs. That is, the hidden state matrix $\textbf{H}$ in Eq. ~\ref{eq:attn} is populated by the pretrained word embeddings.  Fig ~\ref{fig:time_vs_len} shows the  average runtime per epoch (averaged over 10 epochs) of LAMA Encoder (LE) and Transformer Encoder (TE) when sequence lengths are increased from 50 to 250 for IMDB and Reuters datasets. It can be seen from the figure that TE is computationally more expensive per epoch than LE.
LE is also a competitive model with test accuracies of 83.0 \%  on IMDB and 93.9\% on Reuters as compared to 81.7\% (IMDB) and 90.1\% (REUTERS) of TE.


~\subsection{Parameters vs Accuracy}
\begin{figure}
 \centering
   \includegraphics[width=0.45\textwidth]{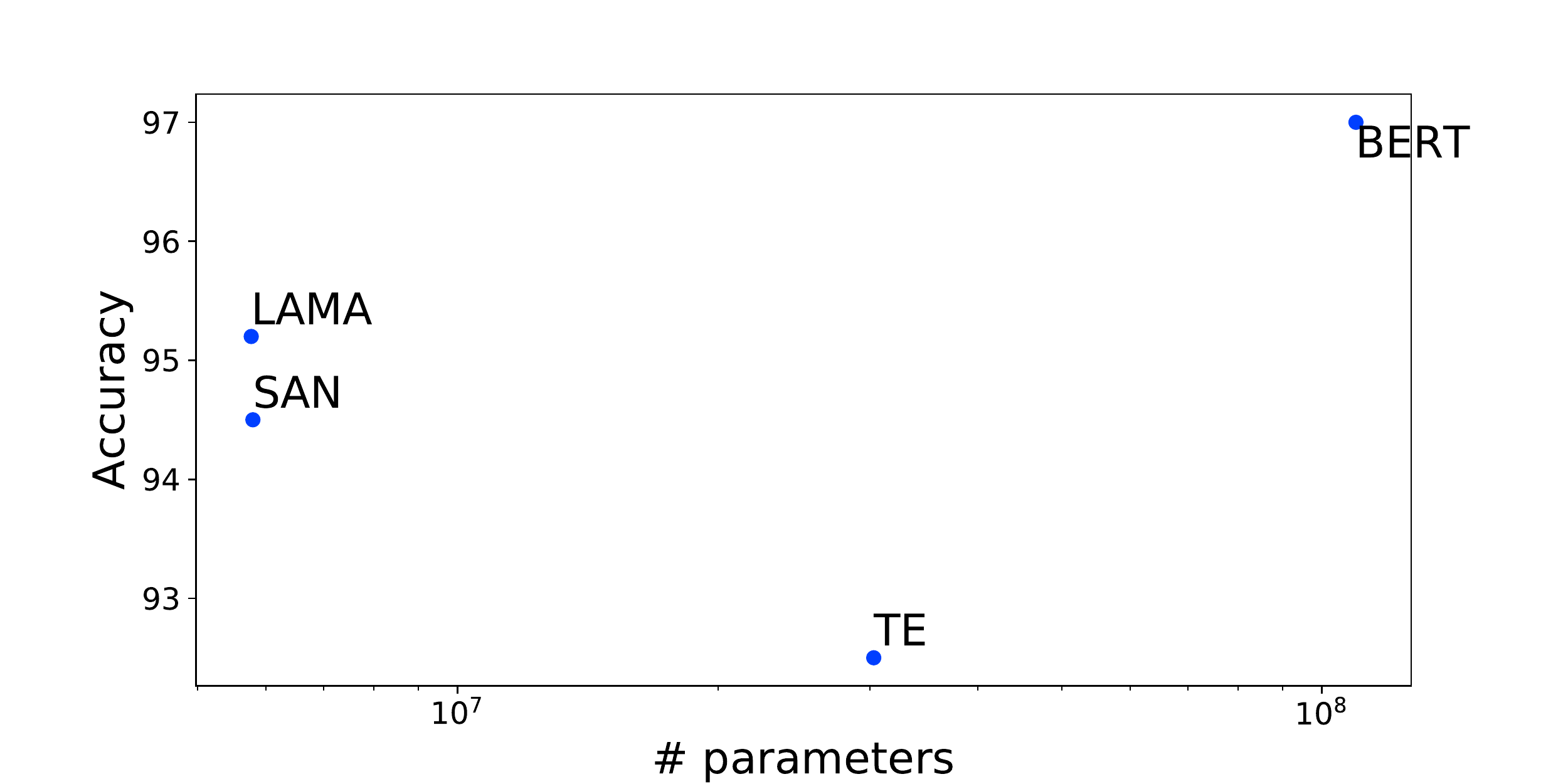}
 \caption{\small{Figure shows the accuracy (y-axis) of the models LAMA, SAN, TE and BERT models on YELP-P dataset when viewed against the model parameters (x-axis). LAMA outperforms SAN and TE while also being more parameter efficient. BERT outperforms LAMA by 1.8\% but by more than an order of magnitude increase in parameters.} }
 \label{fig:perf_vs_params}
 \end{figure}

On the YELP-P dataset our model is outperformed by BERT by 1.8 \%. However it it worth noting the cost of this performance improvement. Fig. ~\ref{fig:perf_vs_params} shows the accuracy of different models and their  corresponding parameters on the YELP-P dataset. It can be seen that \textsf{LAMA+ctx} outperforms SAN \& TE with fewer parameters (difference being linear). BERT slightly outperforms \textsf{LAMA+ctx} however with more than an order of magnitude increase in parameters.

\subsection{Contextual Attention Weights}

\begin{figure}[!t]
 \centering
  \includegraphics[width=0.5\textwidth]{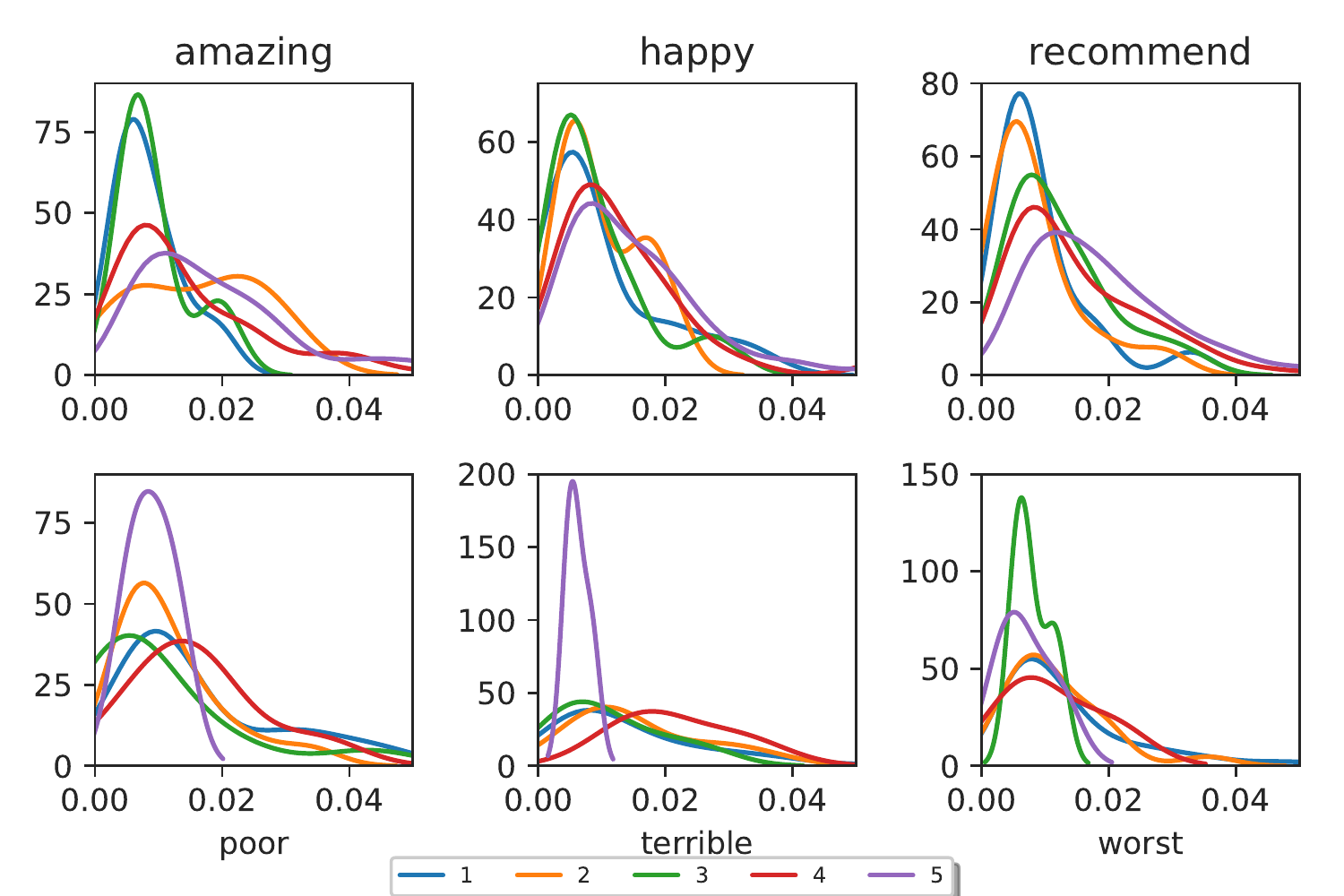} \caption{\small{Attention weight ($x$-$axis$) distribution of the positive words `amazing', `happy' \& `recommend' and negative words `poor', `terrible', \& `worst'. Positive words tend to get higher weights in reviews with higher ratings (3-5) whereas negative words get higher weights for lower ratings (1-2)} Example `terrible' and `poor' get a very low weight for reviews with ratings 5 and `recommend' and `amazing' get high attention weights for reviews with ratings 4 and 5.}
 \label{fig:probs_amazing}
 \end{figure}

To verify that our model captures context dependent word importance and is interpretable we peform qualitative analysis.  We plot the distribution of the attention weights of the positive words `amazing', `happy' and `recommended' and negative words `poor', `terrible' and `worst' from the test split of the Yelp data set as shown in Figure \ref{fig:probs_amazing}. We plot the distribution when conditioned on the ratings of the reviews from 1 to 5. It can be seen from Figure ~\ref{fig:probs_amazing} that the weight of positive words concentrates on the low end in the reviews with rating 1 (blue curve). As the rating increases, the weight distribution shifts to the right. This indicates that positive words play a more important role for reviews with higher ratings. The trend is opposite for the negative words where words with negative connotation have lower attention weights for reviews with rating 5 (purple curve). However, there are a few exceptions. For example, it is intuitive that `amazing' gets a high weight for reviews with high ratings but it also gets a high weight for reviews with rating 2 (orange curve). This is because, inspecting the Yelp dataset we find that `amazing' occurs quite frequently with the word `not' in the reviews with rating 2; `above average but not amazing', `was ok but not amazing'. Our model captures this phrase-level context and assigns similar weights to `not' and `amazing'. `not' being a negative word gets a high weight for lower ratings and hence so does `amazing'. 
Similarly, other exceptions such as `terrible' for rating 4 can be explained due to the fact that customers might dislike one aspect of a business such as their service but like another aspect such as food.

\begin{table}[]
\centering
\caption{\small{Top attended words for Yelp dataset from reviews with ratings 1 and 5 (indicated in paratheses) and Reuters(r8) Dataset.}}
\label{tab:kw}
\begin{tabular}{@{}llll@{}}
\toprule
Yelp(1) & Yelp(5)  & r8(ship)& r8(money) \\ \midrule
\begin{tabular}[c]{@{}l@{}}inconsiderate \\ rudest \\ goodnight \\ worst\\ ever\\ boycott\\ loud\\ livid\\ some\\  hassle\\  friendly \\ ugh \\ dealership\\ brutal \\ rather \\ 1 \\ pizza\\ slow \\ torture \\ absolute\end{tabular} & \begin{tabular}[c]{@{}l@{}}recommend \\ trust\\ referred \\ downside\\  professional\\  100\\  happy\\ attentive\\  stars \\ please \\ delicious \\ safe\\ worth\\ very\\ would \\ blown\\  sensitive \\ removed\\ impressed \\ looks\end{tabular} & \begin{tabular}[c]{@{}l@{}}kuwait \\ gulf \\ south\\ tanker \\ cargo\\ warships \\ pentagon \\ says \\ begin \\ iranian \\ shipping \\ from\\ demand \\ salvage \\ trade \\ production \\ japan \\ gulf \\ india\\  combat\end{tabular} & \begin{tabular}[c]{@{}l@{}}currencies\\ monetary\\ miyazawa \\ stoltenberg\\ accord \\ louvre \\ fed \\ cooperation\\  rate \\ poehl\\ exchange\\ stability \\ currency \\ german \\ reagan \\ pact \\ policy \\support \\trade\\deficit\end{tabular} \\ \bottomrule
\end{tabular}
\end{table}
To further illustrate context-dependent word importance Table ~\ref{tab:kw} lists top attended keywords for Yelp and Reuters datasets. 
Note that superlatives such `100 stars' appear in the list which are strong indicators of the sentiment of a review.

\subsection{Why Multiple Heads}
\label{sec:why_multi}
\begin{figure}[H]
   \includegraphics[width=0.5\textwidth]{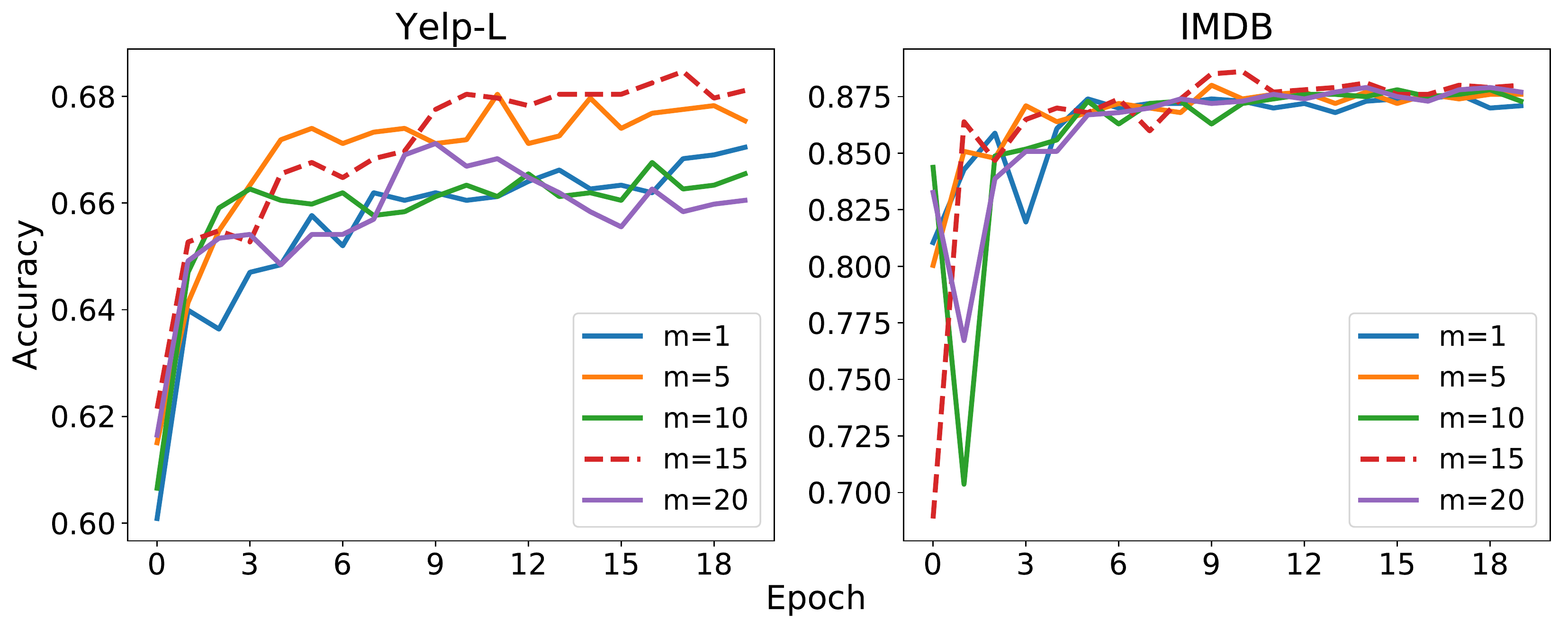}
 \caption{\small{Figure shows the effect of using multiple attention heads. Validation accuracy of \textsf{LAMA+Ctx} is plotted for different values of $m$ for the Yelp-L dataset(left) and the IMDB dataset(right). $x$-$axis$ indicates the number of epochs, $y$-$axis$ indicates the accuracy. Accuracy peaks at $m=15$ for both Yelp-L and IMDB. } }
 \label{fig:acc}
 \end{figure}

Previous works have shown that more heads does not necessarily lead to better performance for machine translation tasks ~\citep{michel2019sixteen}. In this analysis, we seek to answer a similar question for text classification. 

We evaluate the model performance as we vary the number of attention heads $m$ from 1 to 25.  Specifically, we plot the validation accuracy vs. epochs for different values of $m$, for the Yelp-L and IMDB datasets. We vary $m$ from 1 to 20 to get 5 models with $m=1$, $m=5$, $m=10$, $m=15$ and $m=20$. The plots are shown in Figure ~\ref{fig:acc}. From the figure we can see that for the Yelp-L dataset model performance peaks for $m=15$ and then starts falling for $m=20$. We can clearly see a significant difference between $m=1$ and $m=20$, showing that having a multi-aspect attention mechanism helps. For the IMDB dataset model with $m=15$ performs the best whereas model with $m=1$ performs the worst although performances for $m=5,15,20$ are similar.

\section{Conclusion} \label{sec:conclusion}
In this paper we presented a novel compact multi-head attention mechanism and illustrated its effectiveness on text classification benchmarks. The proposed method computes multiple attention distributions over words which leads to contextual sentence representations.
The results showed that this mechanism performed better than several other approaches with fewer parameters. We also verified that the model captured context-dependent word importance. 
We envision two directions for future work -- 1) Currently, the model relies on RNNs that makes it slower due to their sequential computation. We seek to investigate ways of adapting the proposed mechanism in Transformer-style self-supervised language models such as BERT, XLNet etc. without dependency on RNNs by incorporating positional embeddings ~\citep{GehringConvS2S} for faster and efficient learning of language representations; 2) currently, the model uses a single global context vector as the query vector that conflates the entire sentence into one vector which could lead to information loss. Transformer models on the other hand use fine-grained context by querying every word in the sequence for each candidate word. Even though this may help develop direct connections between relevant words it might get redundant. In the future work, we seek to investigate ways of incorporating phrase-level queries as a middle ground between a single global context vector like the proposed approach and fine-grained queries like Transformer providing a balance between complexity and context.


\bibliographystyle{ACM-Reference-Format}
\bibliography{sample-base}

\appendix









\end{document}